\documentclass[runningheads]{llncs}
\usepackage{graphicx}
\usepackage{amsmath,amssymb,amsfonts}
\usepackage{empheq}
\usepackage{algorithm}
\usepackage[noend]{algpseudocode}

\usepackage{graphicx}
\graphicspath{{figures/}}
\usepackage{textcomp}
\usepackage{xcolor}
\usepackage{booktabs}
\usepackage{multirow}

\usepackage{flushend}
\usepackage{amsmath}
\usepackage{bm}
\usepackage{subfigure}
\begin{document}
\title{Robotic Brain Storm Optimization: \\ A Multi-target Collaborative Searching Paradigm for Swarm Robotics \thanks{This work is partially supported by the Science and Technology Innovation Committee Foundation of Shenzhen under the Grant No. JCYJ20200109141235597 and ZDSYS201703031748284, National Science Foundation of China under grant number 61761136008, Shenzhen Peacock Plan under Grant No. KQTD2016112514355531, Program for Guangdong Introducing Innovative and Entrepreneurial Teams under grant number 2017ZT07X386, and Special Funds for the Cultivation of Guangdong College Students Scientific and Technological Innovation (“Climbing Program” Special Funds, PDJH2020b0522).}}
\titlerunning{Robotic Brain Storm Optimization}
%
\author{Jian Yang\inst{1} \and
Donghui Zhao\inst{1} \and
Xinhao Xiang\inst{1} \and Yuhui Shi*\inst{1}}
\authorrunning{J. Yang et al.}
%
\institute{Department Computer Science and Engineering, Southern University of Science and Technology (SUSTech), Shenzhen 518055, China\\
  Correspondence: \email{shiyh@sustech.edu.cn}}
\maketitle              
\begin{abstract}
  Swarm intelligence optimization algorithms can be adopted in swarm robotics for target searching tasks in a 2-D or 3-D space by treating the target signal strength as fitness values. Many current works in the literature have achieved good performance in single-target search problems. However, when there are multiple targets in an environment to be searched, many swarm intelligence-based methods may converge to specific locations prematurely, making it impossible to explore the environment further. The Brain Storm Optimization (BSO) algorithm imitates a group of humans in solving problems collectively. A series of guided searches can finally obtain a relatively optimal solution for particular optimization problems. Furthermore, with a suitable clustering operation, it has better multi-modal optimization performance, i.e., it can find multiple optima in the objective space. By matching the members in a robotic swarm to the individuals in the algorithm under both environments and robots constraints, this paper proposes a BSO-based collaborative searching framework for swarm robotics called Robotic BSO. The simulation results show that the proposed method can simulate the BSO's guided search characteristics and has an excellent prospect for multi-target searching problems for swarm robotics.

  \keywords{Swarm Robotics \and Multi-target Searching \and Brain
    Storm Optimization}
\end{abstract}
\section{Introduction}

Swarm robotics simulates the emergent behaviors of social insects or animals, such as bees, ants, birds, fish schools, wolves, and even humans \cite{yang2018line}. Many works in the literature have shown that this kind of system can be applied to different aspects varies from space exploration to military inspection, from industrial maintenance to medical solutions \cite{yang2016formation,yang2018v}. Collaborative searching problem is one of the popular topics in this field. It intends to determine the specific target(s) location in a particular region in a collective way. It belongs to the coordinated motion and decision making of multi-robot systems. This problem strongly correlates with many practical applications, such as search and rescue applications (scream search, radiation source location, pollution source location, etc.). The characteristics of this problem are that the target number and location(s) are both unknown. Because swarm robot systems have excellent redundancy
based robustness, they have more advantages than single robots or a small group of robots to perform such tasks.

The swarm intelligence optimization algorithms can be applied to target searching tasks of robotic swarms by introducing practical constraints and mapping individuals in an algorithm into robots in a swarm, including the Particle Swarm Optimization (PSO) \cite{yang2019Extended}, Bees Algorithm (BA) \cite{jevtic2010building}, Artificial Bee Colony (ABC) \cite{banharnsakun2012target}, Ant Colony Optimization (ACO) \cite{hoff2010two}, Bacterial Foraging Optimization (BFO) \cite{yang2014target}, Glowworm Swarm Optimization (GSO) \cite{krishnanand2005detection}, Firefly Algorithm (FA) \cite{palmieri2016discrete}, and Grey Wolf Optimizer (GWO) \cite{jain2018Odor}, etc. However, when multiple targets need to be located in an environment, many current solutions imitating swarm intelligence algorithms will converge to specific positions prematurely and lose the ability to explore the searching space further. The multi-target search ability is a critical issue that needs to be solved in the
collaborative searching domain of swarm robotics. What is illustrated in this paper is a new paradigm for collaborative searching of swarm robotics called Robotic Brain Storm Optimization (RBSO). It applies the Brain Storm Optimization (BSO) algorithm \cite{shi2011Optimization} as the source of inspiration and aims to solve the premature problem in the multi-target collaborative searching tasks.

The original BSO algorithm imitates the brainstorming process to solve optimization problems heuristically. It can guide the searching process to converge to optimal solutions over iterations. It has been verified successfully in many real-world applications \cite{yang2020visual}. Individuals in BSO are grouped and diverged in the search space. The search performance could be benefited from this inherent advantage of clustering, making it more suitable for multi-modal optimization problems with multiple peaks. By matching the members in a robotic swarm to the individuals in the algorithm, the BSO has a significant potential to be applied as a new collaboration paradigm for multi-robot systems. Essentially, the multi-target search problem in swarm robotics can be transformed into a multi-modal optimization problem in a two-dimensional or three-dimensional space \cite{yang2020exploration}. The BSO's multi-modal optimization properties make it an excellent candidate to solve the collaborative multi-target
searching problems. 
It worth to mention that there are related applications of swarm robotics that use the SI algorithms as optimization tools. For example, the Brain Storm Robotics (BSR) framework can automatically design the corresponding swarm behavior collectively \cite{yang2020brain}. Li et al. expressed swarm robotics exploration as an optimization problem and then used the BSO method to solve it \cite{li2019unknown}. There are three kind of operations in the original BSO algorithm: clustering, new individuals generation, and selection, as shown in Alg.\ref{alg:bso}.  \vspace{-1em}
\begin{algorithm}[!htb]
  \caption{The BSO Procedure} \label{alg:bso}
  \begin{algorithmic}[1]
    \State Randomly generate $n$ potential solutions (individuals);

    \State Evaluate the generated $n$ solutions;

    \While{not terminated}

    \State \textbf{Clustering}: Cluster $n$ individuals into $m$ clusters by a clustering algorithm;
    
    \State \textbf{New individuals generation}: randomly select one or two cluster(s) to generate $n$ new individuals;
    
    \State \textbf{Selection}: The fitness values of the newly generated individuals are compared with the existing individuals with the same index, the better one is kept;
    
    \EndWhile
  \end{algorithmic}
\end{algorithm}

The purpose of clustering in the solution space is to converge the solution to a smaller area \cite{cheng2017comprehensive}. 
Then, a new individual can be generated based on one or several cluster(s). In the original BSO, a probability value $p_{one}$ is adopted to determine while a new individual will be generated by one or two cluster(s). Generating an individual from a cluster can refine the search area and improve exploitation capabilities. Conversely, individuals generated from two or more clusters may be far away from these clusters. In this case, the exploration ability is enhanced. Furthermore, the BSO uses another predefined probability value $p_{center}$ to determine whether to generate a new solution based on the cluster center(s) or non-cluster center(s). In the one cluster generation case, the new individual from center or normal individual can control the exploitation region. While in several clusters generation case, the normal individuals could increase the population diversity of swarm. The selection strategy in BSO is to keep good solutions in all individuals. The better solution is kept by the selection strategy after each new individual generation, while clustering strategy and generation strategy add new solutions into the swarm to keep the diversity for the whole population.

The rest of the paper is organized as follows: 
Section 2 defines the collaborative multi-target searching problem and the assumptions for this paper. Section 3 introduces the RBSO framework, including operations of grouping, new position generation, task allocation, fitness evaluation, etc. The simulation results with the proposed framework are given in Section 4. 
The conclusion is reached in Section 5 with the forecast of our future works.

\section{Problem Statement}
\subsection{Assumptions}
\subsubsection{Environments and Targets}
For the sake of simplicity, we intend to conduct related research in
this article in a two-dimensional environment. This work can also be
further extended to three-dimensional environments. In this
hypothetical two-dimensional environment, some simulated obstacles
will be randomly distributed. During the movement, the robot must
avoid these obstacles, and the robots cannot collide
either. Meanwhile, a number of static targets will be distributed in
the same environment outside the obstacles. The targets can broadcast
non-directional beacon signals. Futhermore, it can only cover a small
area around it, and the signal strength will attenuate with distance,
as shown in the following formula:
\begin{equation}
  \label{eq:attu}
  s = \frac{1}{a\sqrt{\pi}}e^{{-d}/{a^2}}
\end{equation}
where $s$ is the signal strength, $d$ is the distance to the center position of the target, $a$ is the attenuation coefficient. 

\subsubsection{Member Robots}
Simple but without generality loss, this article uses a homogenous
swarm with the omnidirectional model for member robots. Furthermore,
in terms of perception, we assume that every member robot is equipped
with a sensor that can detect the target signal strength, i.e., it can
measure the target signal strength at the current position. Also, it
can perceive other robots and obstacles within a specific range. In
terms of positioning, it is assumed that the member robot can obtain
its position in the reference coordinate system and knows the boundary
of the search area. In terms of communication, it is assumed that
information can be shared with all other robots through corresponding
information interaction within a time slot. In terms of storage, it is
assumed that member robots have storage capacity and can record the
target signal strength of the visited location.

\subsubsection{Target Handling}
For target handling, we assume that when the detected target signal is
stronger than a certain threshold, which can be converted into a
distance less than $\epsilon$ to the target, the target is treated as
found. Before the next iteration, the robot will stay to process the
target until the next iteration. After the handling, the target will
no longer broadcast beacon signals to the environment, and the robot
that processes the target will become available again.

\subsection{Multi-target Searching Problem}
Generally, based on the number of targets in a given search area, collaborative search tasks can be divided into the following two categories: single-target search and multi-target search. If the target is moving, the problem is a dynamic single-target or multi-target search problem correspondingly. This article aims to design a method of controlling a swarm of mobile robots, so that the path of the robots will contain the position of the targets as many as possible, i.e., to locate the multi-target positions. Denote $m$ is the target number in an unknown aera, $L_m$ is the location of target $m$, $\epsilon$ is the allowed tolerance vector, $P_i(t)$ is the position of robot $i$ at time $t$, the multi-target collaborative searching problem of swarm robotics can be formally expressed by Eq.(\ref{eq:card}):
\begin{equation}
  \label{eq:card}
  \begin{aligned}
    & \max_{P_i} |\{m \in \mathbb{N}|L_m \pm \bm{\epsilon} \in \sum_{t=0}^T\sum_{i=1}^NP_i(t)\}| \\
    s.t. &\qquad P_i(t) \neq P_j(t) \quad  \forall i,j \in N, i \neq j \\
    & \qquad P_i(t) \neq P_o(t) \quad \forall i \in N, o \in O
  \end{aligned}
\end{equation}
where $t$ and $T$ is the current and maximum searching time respectively, $N$ is the population of the robotic swarm. $|\cdot|$ is the cardinal number of the set. $P_i(t) \neq P_j(t)$ represents the anti-collision between robots, and $P_i(t) \neq P_o(t)$ means the robobts in the swarm need to avoid obstacles in environments, which expressed by set $O$ in Eq.(\ref{eq:card}). It should be noted that although we have formalized the multi-object search problem of swarm robotics above in an optimization form, it can not be solved directly by optimization mathmatically. The expression contains physical constraints, such as the relationship between the robot's moving speed and its position, collision avoidance, etc., which needs to be achieved in the physical world by robotic techniques.

\section{Robotic Brain Storm Optimization}
The basic idea of this paper is to map the BSO to Robotic BSO. As shown in Table \ref{tab:mapping}. The target searching environments correspond to the BSO algorithm's solution space. The signal strength is treated as the fitness value at a position, and the swarm members are mapped to the individuals in the algorithm. The decision of the robotic swarm is imitating the iterative process of the BSO algorithm. Furthermore, the operations such as the clustering, new individual generation can be mapped to position grouping, new position generation, correspondingly. As modeled above, the multi-target searching problem is related to a multi-modal optimization problem.

\begin{table}[!htb]
  \caption{Mapping BSO Algorithm to RBSO.}
  \label{tab:mapping}
  \centering
  \label{tab:mapping}
  \begin{tabular}{|c|c|}
\hline
BSO                       & RBSO                       \\ \hline
N-D Solution Space        & 2-D or 3-D Searching Space \\ \hline
Fitness Value             & Signal Strength            \\ \hline
Individuals               & Robots                     \\ \hline
Clustering                & Grouping                   \\ \hline
New individual generation & New position generation    \\ \hline
Multi-modal Optimization  & Multi-target Searching     \\ \hline
\end{tabular}
\end{table}

\begin{figure}[!htb]
  \centering \includegraphics[width=0.6\textwidth]{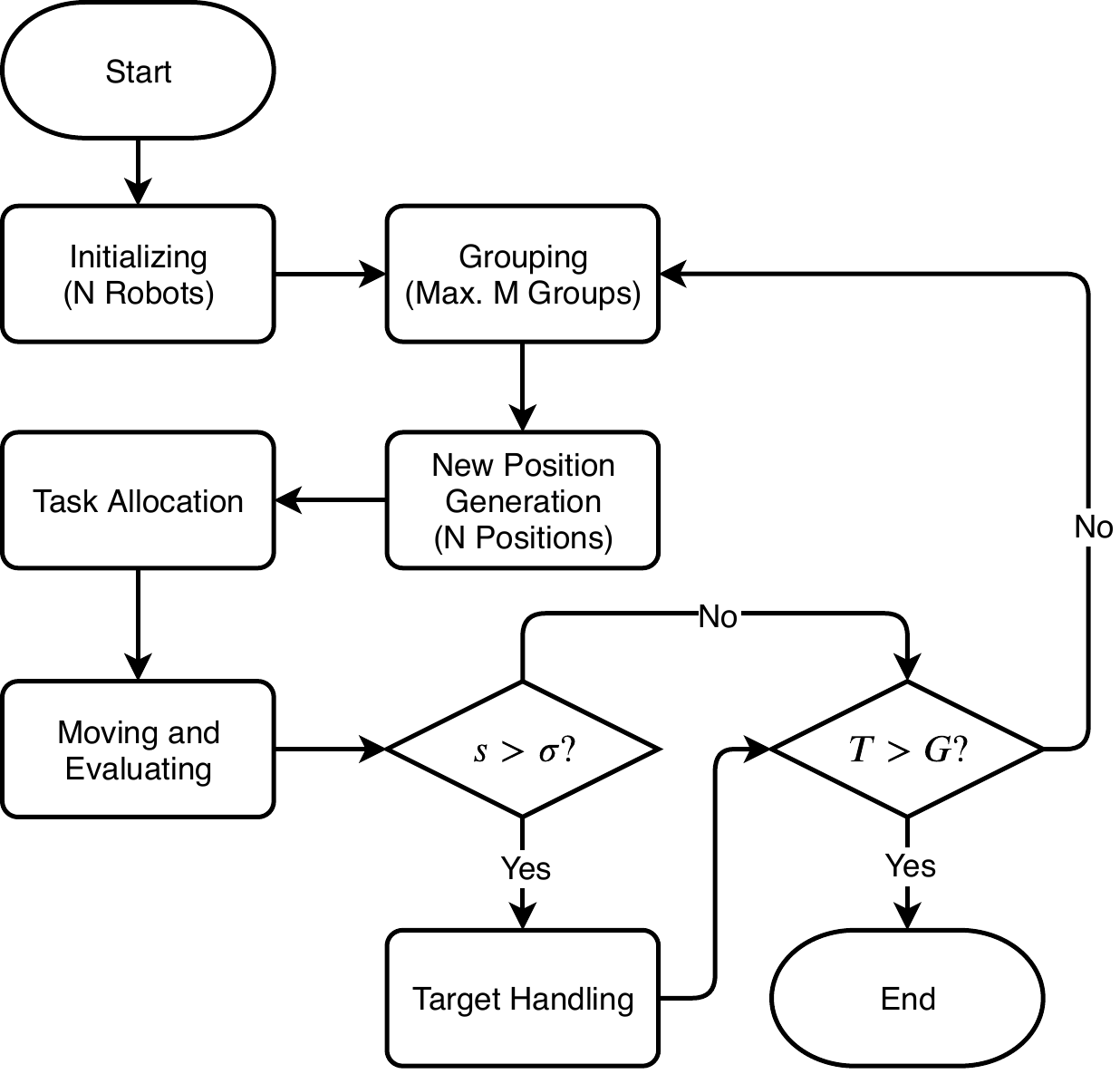}
  \caption{Flowchart of RBSO.}
  \label{fig:flow}
\end{figure}

As Fig.\ref{fig:flow} shows, the proposed method mainly includes the following steps. First, grouping a swarm of robots 
into several subgroups according to their fitness values in their current locations, i.e., the clustering procedure in the BSO algorithm. Secondly, according to the subgroups, a series of new positions are generated, i.e., the new solution generation procedure in the BSO algorithm. Unlike the update method in the BSO algorithm, the fitness values of the newly generated positions can not be obtained immediately. Instead, it has to be physically visited to get the fitness values by the robots. Thus we designed an additional task allocation procedure, which optimally assigns the newly generated positions to each robot in the swarm, then controls the robots move to the assigned positions.
\vspace{-1em}
\vspace{-1em}

Furthermore, during the movements to a generated goal point, the robot can measure the signal strength along the moving path. Therefore, different from the evaluation of a single point by the BSO algorithm, in the proposed RBSO, when evaluating the signal strength of a newly generated position, it evaluates the signal strength of all points on the path between the robot and the allocated position. If the target signal value in a position exceeds the predetermined threshold ($\sigma$) during the movements, it will be marked as a potential target position. As assumed, the robot will stay there to handle the target until the next planning procedure. After all robots in the swarm stop moving, the procedure will return to the grouping and new location generation operations until all the targets have been found or the specified search time has been reached.
\subsection{Grouping}
Corresponding to the clustering operation in the BSO algorithm, the purpose of grouping is to allow 
each robot in a swarm to be divided into several subgroups in the search space. In this article, we simply use a top-down hierarchical clustering method to complete the task of robot grouping, namely DIvisive ANAlysis Clustering (DIANA), which constructs the hierarchy in the inverse order \cite{kaufman2009finding}. Initially, all robots are seen as in the same group, and the largest group is split until the one of the following termination condition is met, i.e. the mean distance between elements of each group ($\bar{D}$) larger than a threshold ($m_d$), or the number of groups reach a pre-defined number $m_g$. In order to ensure that the new solution can be generated according to the BSO mechanism, here $m_g\leq 2$.
\begin{equation}
  \label{eq:grouping1}
  \bar{D} = \frac{1}{|\mathcal{A}|\cdot|\mathcal{B}|}\sum_{x\in \mathcal{A}}\sum_{y\in \mathcal{B}}d(x,y)
\end{equation}
where $\mathcal{A}$, $\mathcal{B}$ is two groups, $d(x,y)$ is the Euclidean distance between two elements $x$ in group $\mathcal{A}$ and $y$ in group $\mathcal{B}$.
\vspace{-1em}
\begin{algorithm}[!htb]
  \caption{Grouping Precedure in RBSO} \label{alg:grouping}
  \begin{algorithmic}[1]

    \State Inputs: $N$ $pbest$, maximum groups $m_g$, maximum iterations $T_g$, $m_d$;

    \State Consider all robots as a whole group;

    \State Selected Group $\mathcal{G}_s$ = the whole group;
    
    \While {not terminated}
    
    \State Find the most dissimilar pairs of robots $(i, j)$ in $\mathcal{G}_s$;
    
    \State Assign the robots closer to $i$ to a new group 1 and the robots closer to $j$ to a new group 2.

    \State Split the $\mathcal{G}_s$ to group 1 and group 2.

    \State $\mathcal{G}_s$ = Find the group with the maximum internal distance in all groups.

    \EndWhile

    \If {$|\mathcal{G}_s| \leq 2$}
    
    \Return

    \EndIf

  \end{algorithmic}
\end{algorithm}
\vspace{-1em}

\subsection{New Position Generation}
After grouping, the new positions can be generated according to the procedure of the BSO. The difference is that the class center is selected according to the fitness values in the BSO, i.e., individuals with a larger fitness value in the group will be selected as the class center. In the collaborative search task of a swarm of robots, since the targets' influence ranges are not global, all the fitness values of $pbest$ of robots in a group maybe 0. Also, there is a case that more than one robots reaches the same maximum fitness value in a group. In these cases, the group center will be randomly selected among the individuals in a group.

\subsection{Task Allocation}
Multi-robot task allocation (MRTA) is an essential aspect of many multi-robot systems. It is a problem of determining which robots should execute which tasks to achieve overall system goals. The features and complexity of the MRTA problem depend on which requirements are under consideration \cite{korsah2013comprehensivea}. In this paper, the task allocation aims to match the newly generated positions to each robot in the swarm for a new round of fitness evaluation.


In this article, the member robot can be regarded as a single-task robot (ST) since it only needs to evaluate the signal strength or handling targets at a time during the searching process. Besides, only one robot is needed for the signal strength evaluation task of a specific position, i.e., the task is a single robot task (SR). Also, the corresponding task allocation is performed after each step of the new position is generated, i.e., the instantaneous allocation (IA). Therefore, the task allocation problem is a typical ST-SR-IA allocation problem \cite{gerkey2004formal}, which is 
an instance of the optimal assignment problem (OAP) from the field of combinatorial optimization . Given $n$ robots and $m$ tasks, each task requiring one worker. The OAP can be cast in many ways, including as an integral linear program that find $mn$ nonnegative integers $\alpha_{ij}$ that minimize the cost of the system:
\begin{equation}
  \label{eq:ta}
  \min C = \sum_{i=1}^n\sum_{j=1}^m\alpha_{ij}c_{ij}
\end{equation}
subject to:
\begin{equation}
  \label{eq:ta2}
  \begin{aligned}
    &\sum_{i=1}^n\alpha_{ij}=1, \qquad 1 \leq j \leq m  \\
    &\sum_{j=1}^m\alpha_{ij}=1, \qquad 1 \leq i \leq n 
  \end{aligned}
\end{equation}
where $C$ in Eq.(\ref{eq:ta}) is the overall system cost, which is the weighted sum of each assignment costs. Eq.(\ref{eq:ta2}) enforces the constraints of single-robot tasks and single-task robots. The $\alpha_{ij}$ are integers that must all be either 0 or 1. This problem is also known as zero-one type integer linear programming. Here the robot number is equal to the generated new locations, i.e., $|m| = |n|$. The linear assignment problem can be solved in polynomial time with algorithms such as the Hungarian algorithm \cite{korsah2013comprehensivea}. 

\subsection{Moving and Evaluating}
After allocating the tasks, each robot corresponds to a newly generated position in the environment. They will move to the assigned position and evaluate the fitness values along the path. Due to obstacles and other robots in the environment, the robot needs a suitable motion planning method to move to the target point as much as possible. 
Here we use a relatively simple modified Bug Algorithm to achieve the corresponding motion planning \cite{zohaib2013iba}. 
When encountering obstacles, the robot will move along the edges of the obstacles. 
The evaluating procedure for each robot is as shown in Algorithm \ref{alg:eval}. During the process of a member robot moves towards the target, it will evaluate the target signal strength at the current position. As mentioned earlier, if the signal strength of the current position is higher than the pre-defined threshold, it will enter the target handling operations. After the target is processed, the will continue to participate in subsequent tasks. 
To avoid the uncertainty caused by individual robots' motion and possible failures, and to ensure the searching process, two termination conditions are set here: one is for all robots to reach the assigned positions, the other is the moving steps reach a maximum movement step $m_s$.
\vspace{-1em}
\begin{algorithm}[!htb]
  \caption{Signal Evaluating Precedure in RBSO} \label{alg:eval}
  \begin{algorithmic}[1]

    \State Inputs: Goal position and $pbest$ for each robot, maximum moving steps $m_s$;

    \ForAll {robot in the swarm}
    
    \While {not terminated}
    
    \State Move forward one step to the goal;

    \State Evaluate the target strength of current position;

    \EndWhile

    \If {$s>\sigma$}

    \State Target Handling;

    \EndIf

    \EndFor

  \end{algorithmic}
\end{algorithm}
\vspace{-2em}

\section{Results}
\begin{figure}[!htb]
  \centering
  \subfigure[Configurations]{\label{fig:configuration} \includegraphics[trim=0 130 0 150,clip,width=0.45\textwidth]{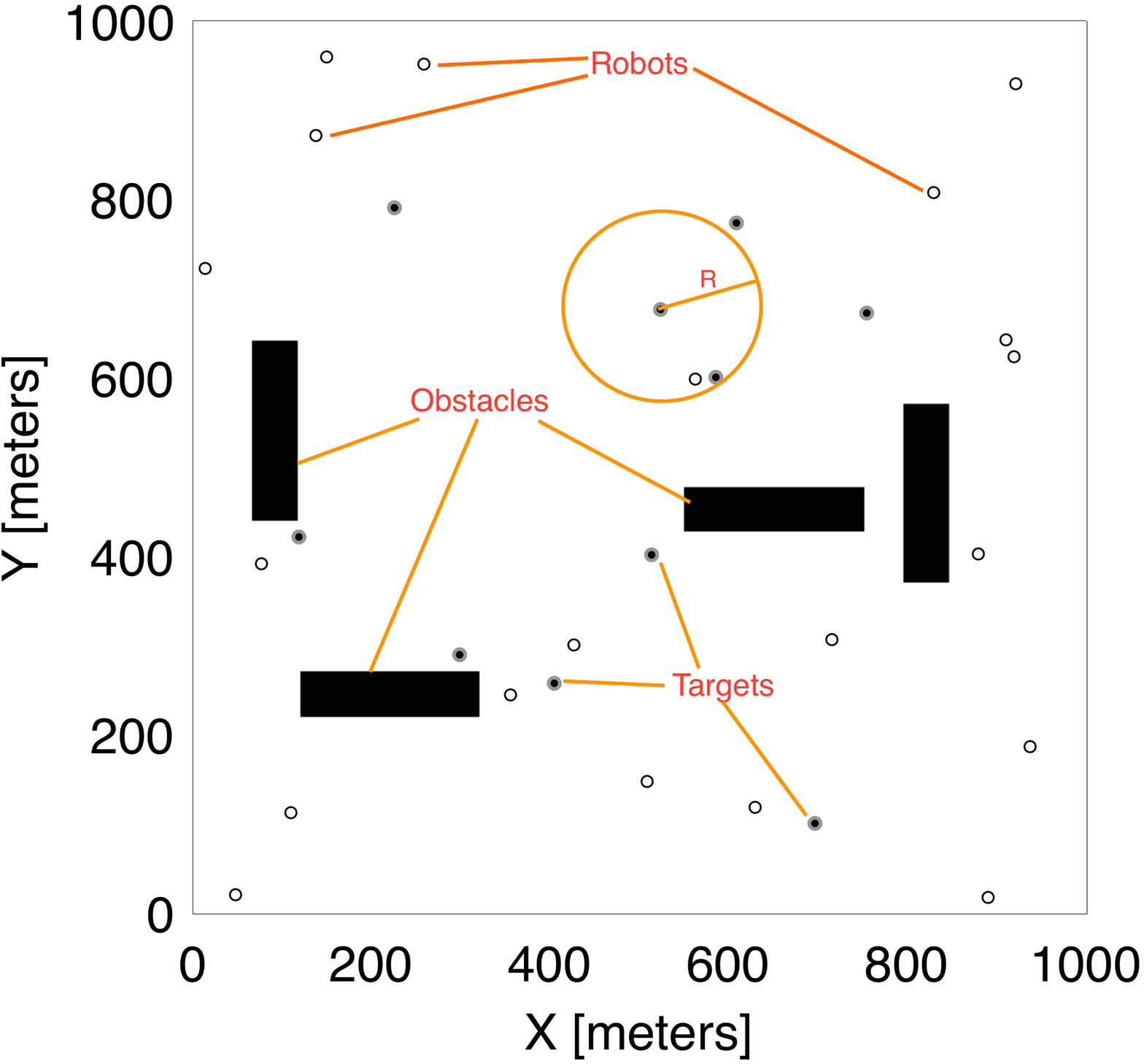}}
  \subfigure[Target Signals]{\label{fig:signals} \includegraphics[trim=120 260 150 250,clip,width=0.45\textwidth]{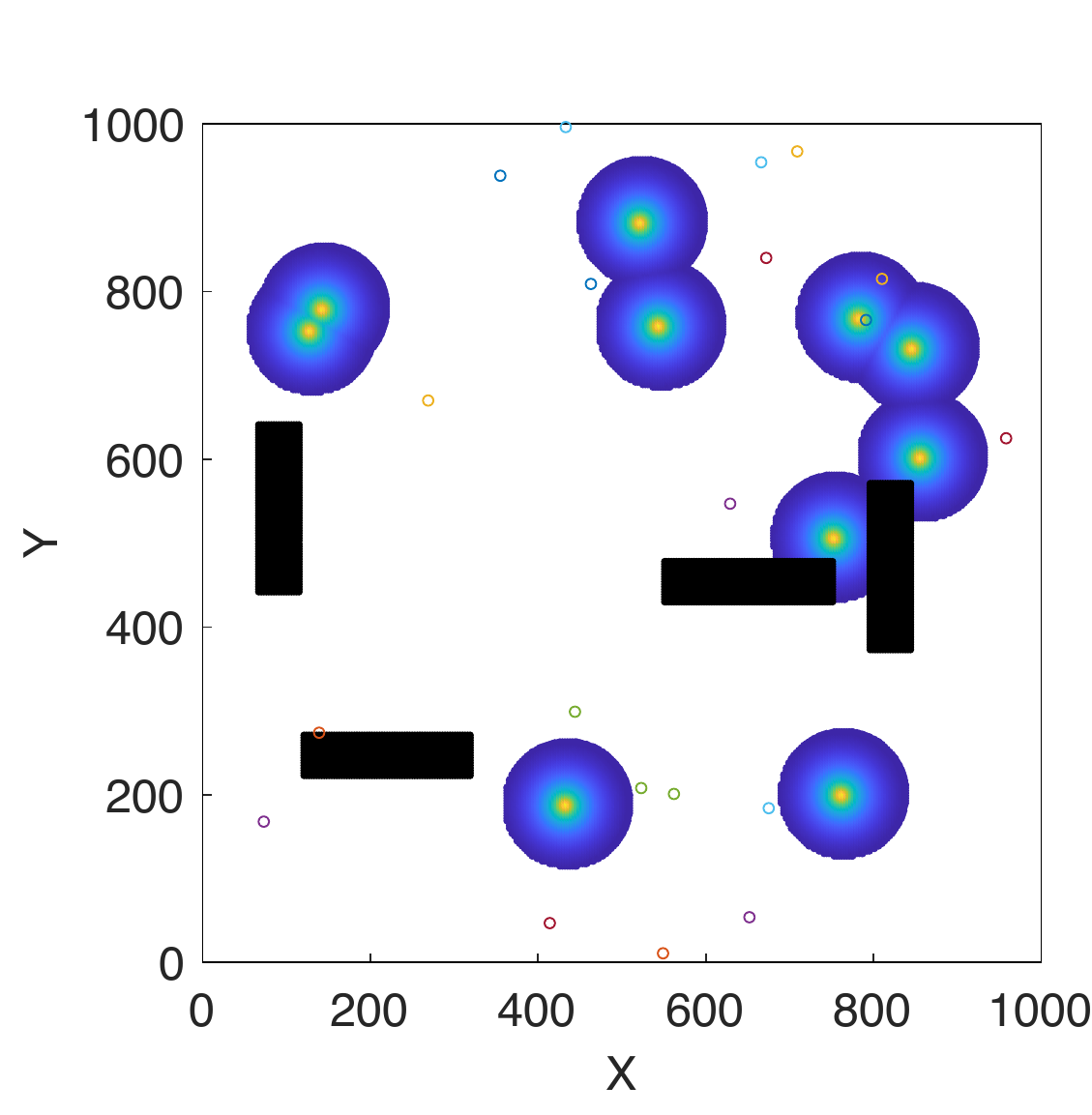}}
  \caption{The simulation environment.}
  \label{fig:config}
\end{figure}

Using the above method, we have obtained some preliminary results. The simulation environment is as shown in Fig.\ref{fig:config}, where 20 member robots and 10 targets are randomly distributed in a 1000$\times$1000 2-D environment with obstacles. The member robots are represented by the hollow circles, the targets are represented by the solid circles, and the obstacles are shown as black rectangles. The range of influence of the target signal is marked with $R$. The signal influence region is shown in \ref{fig:signals}.
\vspace{-2em}
\vspace{-2em}

\begin{figure}[!htb]
  \centering
  \subfigure[T=1]{\label{fig:r1} \includegraphics[trim=50 200 80 220,clip,width=0.31\textwidth]{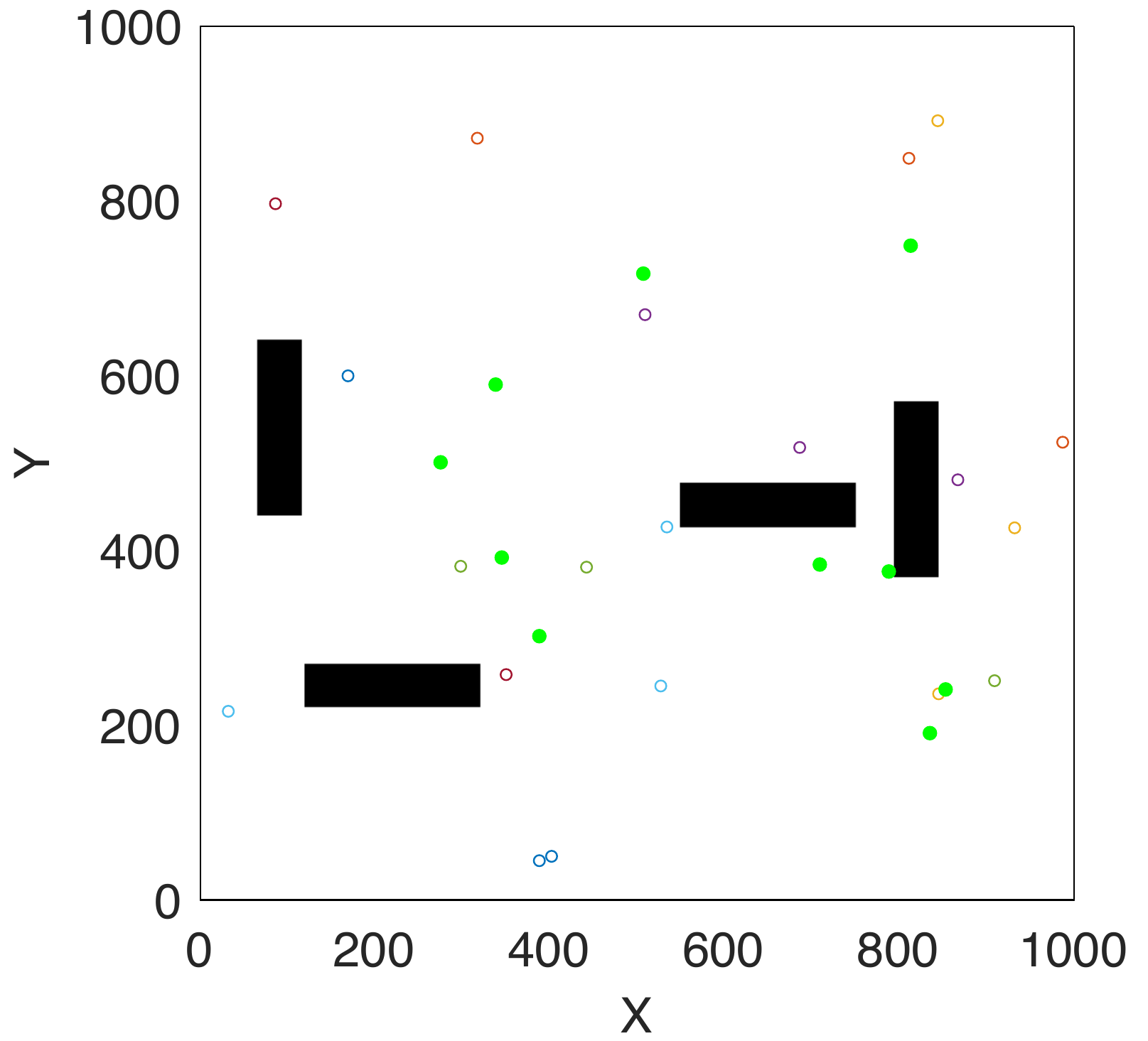}}
  \subfigure[T=400]{\label{fig:r2} \includegraphics[trim=50 200 80 220,clip,width=0.31\textwidth]{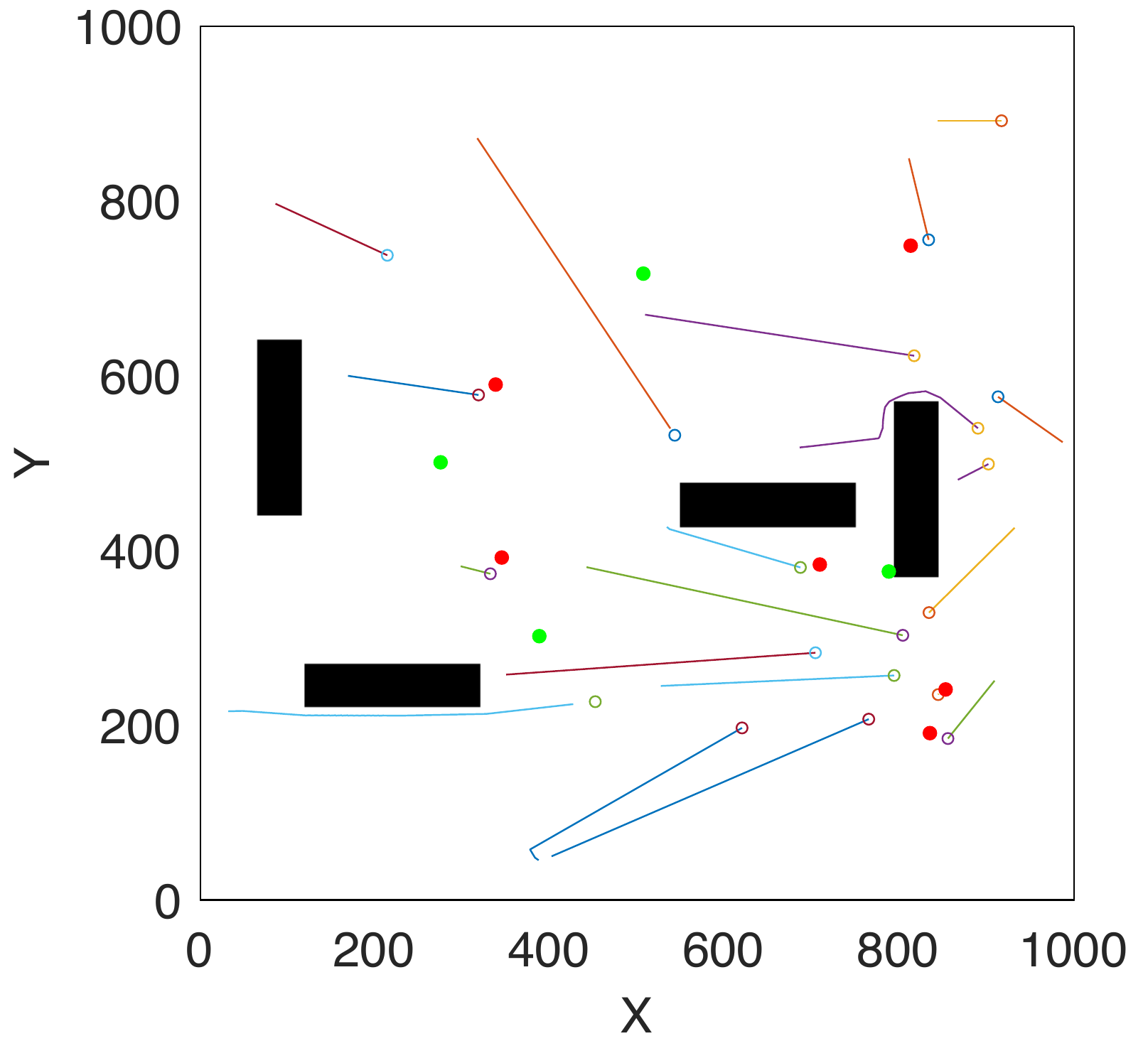}}
  \subfigure[T=1000]{\label{fig:r3} \includegraphics[trim=50 200 80 220,clip,width=0.31\textwidth]{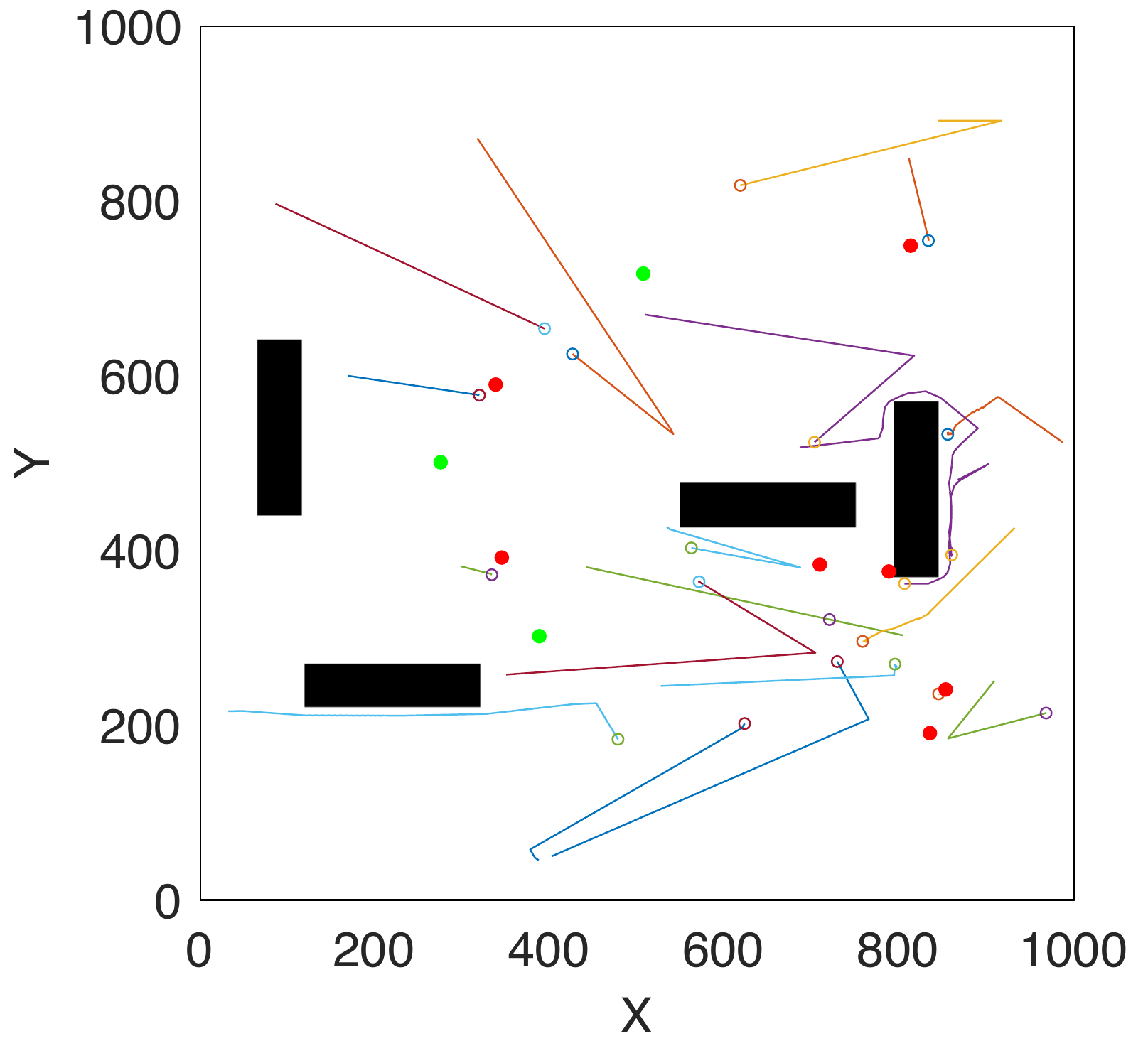}}
  \subfigure[T=2500]{\label{fig:r4} \includegraphics[trim=50 200 80 220,clip,width=0.31\textwidth]{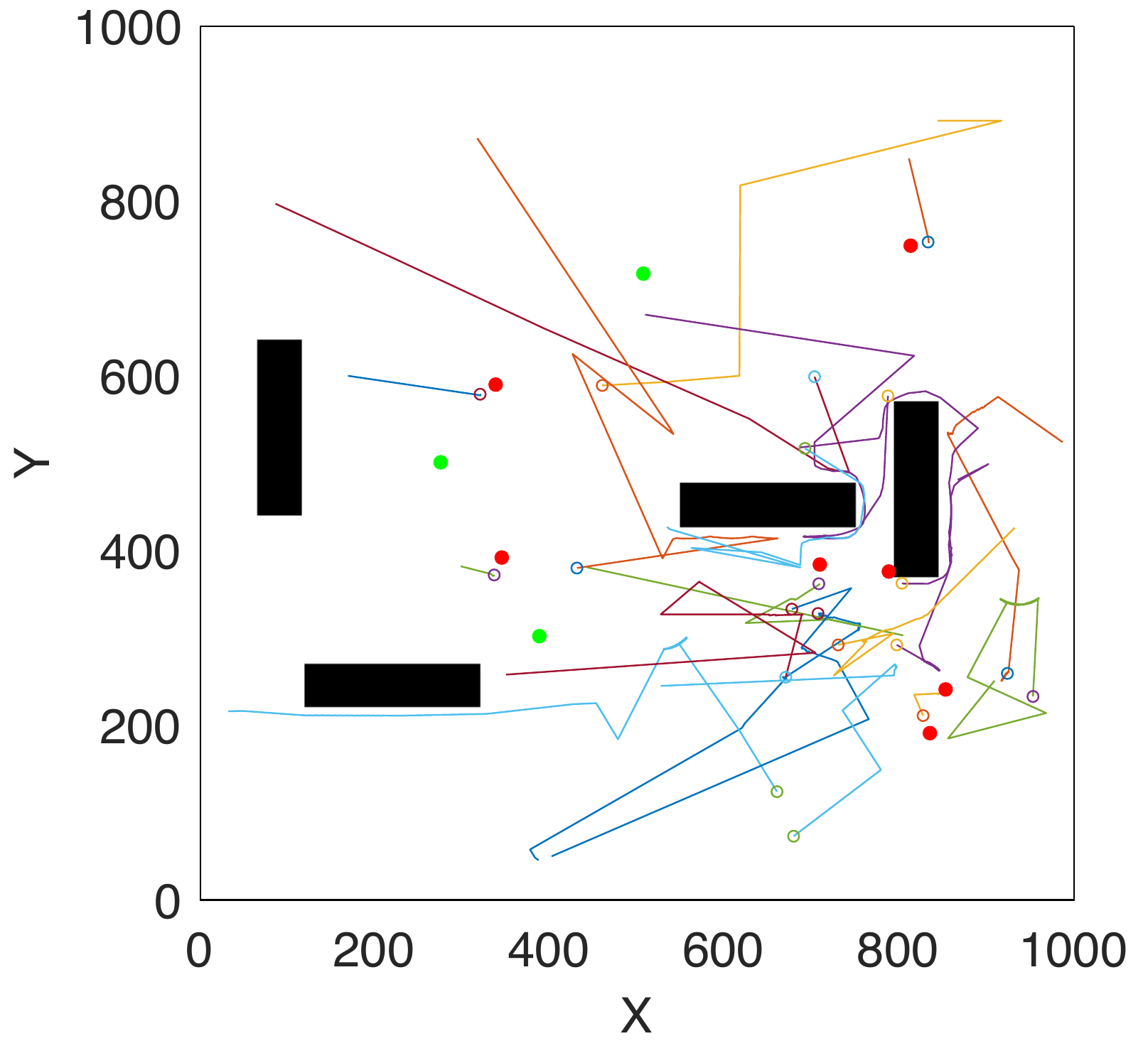}}
  \subfigure[T=4000]{\label{fig:r5} \includegraphics[trim=50 200 80 220,clip,width=0.31\textwidth]{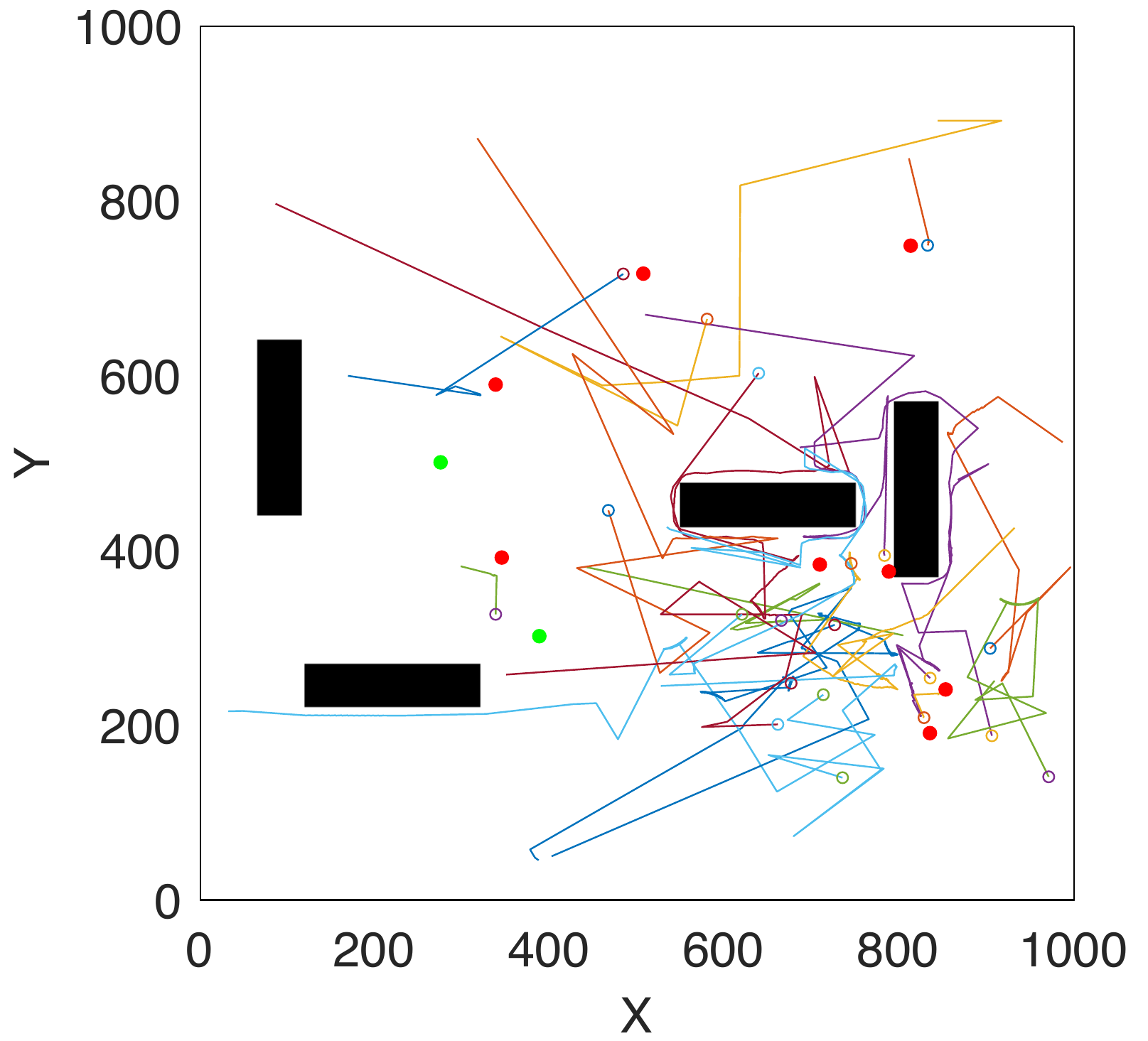}}
  \subfigure[T=5000]{\label{fig:r6} \includegraphics[trim=50 200 80 220,clip,width=0.31\textwidth]{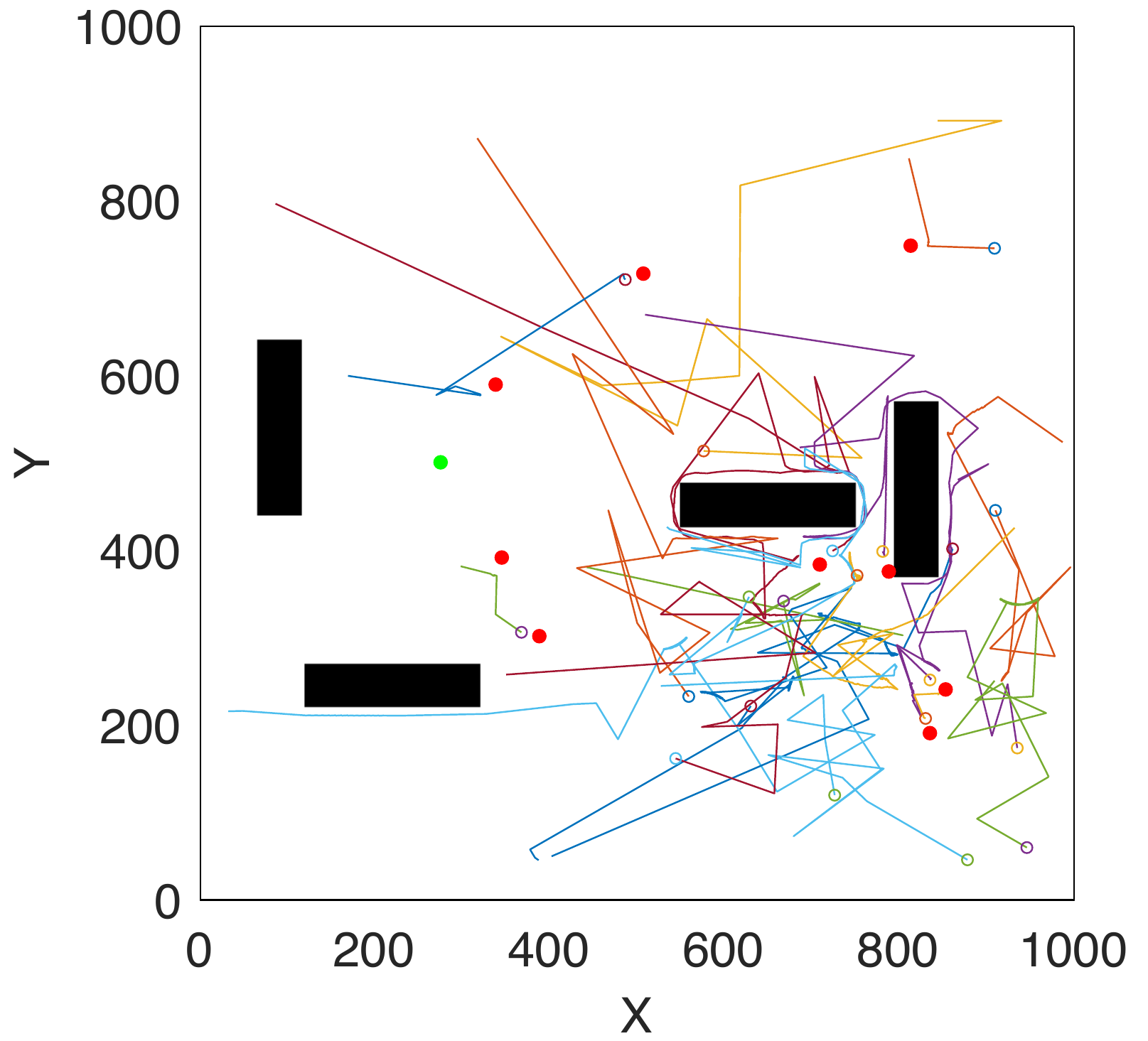}}
  \subfigure[T=6000]{\label{fig:r7} \includegraphics[trim=50 200 80 220,clip,width=0.31\textwidth]{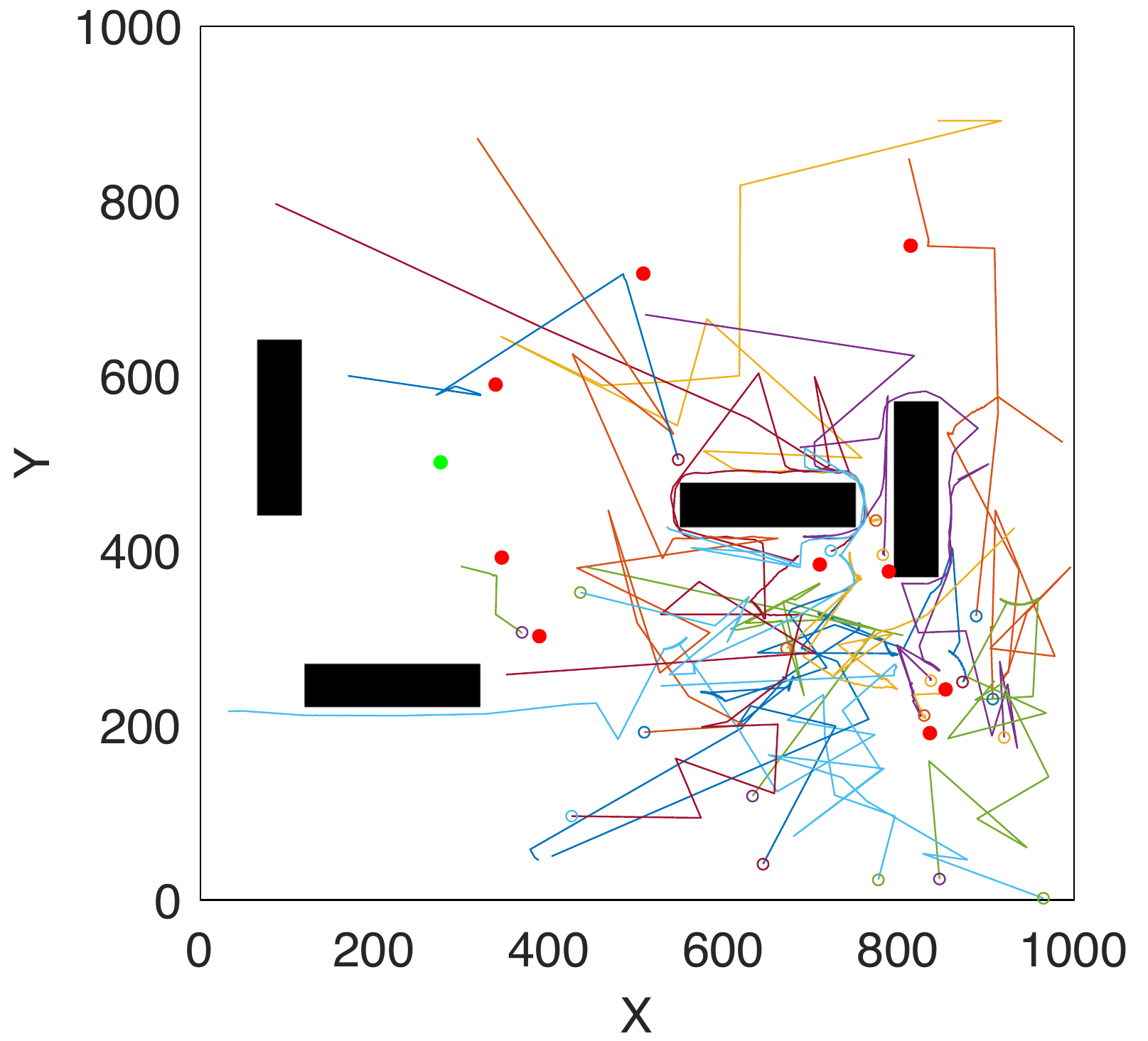}}
   \subfigure[T=8000]{\label{fig:r8} \includegraphics[trim=50 200 80 220,clip,width=0.31\textwidth]{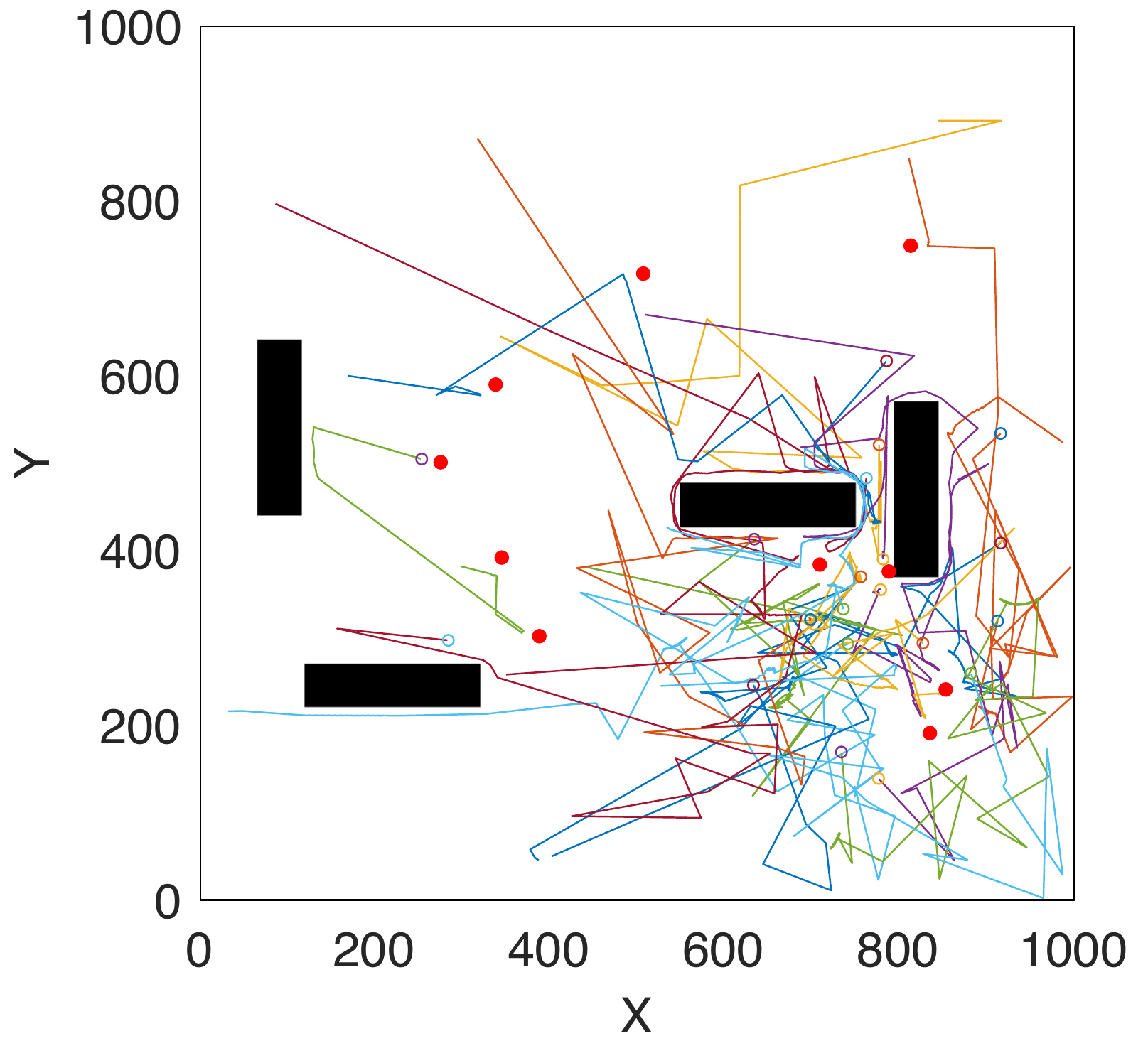}}
  \caption{The simulation results}
  \label{fig:results}
  \vspace{-1em}
\end{figure}

The parameters configurations is shown in Table \ref{tab:config}, where $p_{one}$ and $p_{center}$ are parameters in BSO algortihms for new positon generation. Other parameters such as signal attenuation coefficient $a$, maximum groups $m_g$, maximum iteration generations $T_g$, maximum inter-group mean distance $m_d$, and maximum moving steps $m_s$ for each round of evaluation are RBSO parameters. The sampling time for each moving step is set to 0.1s. The simulations are conducted with mobile robot toolbox in Matlab 2020a on an iMac with 3.6 GHz Intel Core i9, 40GB DDR4 memory.
  \vspace{-1em}
\begin{table}[!htb]
  \centering
  \caption{Simulation Configurations}
  \label{tab:config}
\begin{tabular}{|c|c|c|c|c|c|c|}
\hline
\multicolumn{2}{|c|}{BSO Parameters} & \multicolumn{5}{c|}{RBSO Parameters}        \\ \hline
$p_{one}$       & $p_{center}$       & $a$ & $m_g$         & $T_g$ & $m_d$ & $m_s$ \\ \hline
0.4             & 0.8                & 10  & $\frac{N}{4}$ & 20000 & 250   & 500   \\ \hline
\end{tabular}
\end{table}
  \vspace{-1em}
The preliminary simulated results is shown in Fig.\ref{fig:results}, where he hollow circle represents the robot, and the solid circle represents the target to be searched. When it is not found, it is marked in green and displayed in red after handling. The solid line in the figures are pathes of the member robots.
  \vspace{-1em}

Fig.\ref{fig:r1} shows the initialization distribution of robots and targets, the green dots are the targets to be searched. It will be marked in red when reached. The followed Fig.\ref{fig:r2}-Fig.\ref{fig:r8} show the iterative search process. It can be seen that after a certain number of iterations, 10 targets in the environment can be all found. Since the robot needs to arrive at the newly generated location or reach the targets before a stop, the paths in the graph are in polygonal lines. We further test massive times, the proposed method can find all the targets under the above configurations. The proposed method maps the clustering operation of the original BSO to the grouping operation, which produces the same utility as the original algorithm, i.e., to ensure the convergence of the search process. Besides, the operation of new location generation not only considers the members with the largest fitness value in the group but also according to the members with smaller fitness value with a certain probability, which ensures diversity of the searching process. Furthermore, since each generated position needs to be accessed by a robot, the newly added task allocation module allocates the generated positions to each robot in an optimal way without conflicts. The above results indicate the effectiveness of the proposed method, which is worthy of further development.




\section{Conclusion}
In this paper, the Robotic Brain Storm Optimization (RBSO) for cooperative search tasks of swarm robotics is proposed. In this method, individuals in the BSO algorithm are mapped to the members in robotics swarms. Relevant constraints are introduced for the multi-target cooperative search task of swarm robot systems. The proposed method imitates the optimal searching process of the original BSO algorithm with the considerations of robot physical limits and motion constraints. In the BSO optimization algorithm, the position update of the individuals in the solution space is neither restricted by physical limitations such as movement velocity nor restricted by requirements such as collision avoidance. The proposed paradigm takes both of the above into consideration. The preliminary results show that the proposed method inherits the multi-mode optimization performance of the BSO algorithm, which can be applied in multi-target searching problems. In the future, we will further evaluate its statistical performance and perform more comparative research with other methods.

%
%
%
\bibliographystyle{splncs04}
\bibliography{RBSO1}

\end{document}